\documentclass[11pt]{article}
\usepackage{kotex}
\usepackage[final]{acl}

\usepackage{times}
\usepackage{latexsym}
\usepackage{amssymb}
\usepackage{graphicx} 
\usepackage{booktabs}
\usepackage{placeins}

\usepackage[T1]{fontenc}

\usepackage[utf8]{inputenc}

\usepackage{microtype}

\usepackage{inconsolata}

\usepackage{graphicx}

\title{Structurally Different, Functionally Persistent: Capability Regions Under Model Optimization}

\author{
  \textbf{Jiwoo Kum, Chanwoo Kim, Doyun Kim, Seonghyeon Cho, Sihyeon Lee, Chanjun Park\textsuperscript{$\dagger$}} \\
  Soongsil University \\
  \texttt{\{goldi1204, kcw4776, doyunkim, chosh040, sihyeon2988\}@soongsil.ac.kr} \\
  \texttt{chanjun.park@ssu.ac.kr}
}

\begin{document}
\maketitle

\begingroup
\renewcommand{\thefootnote}{\fnsymbol{footnote}}
\footnotetext[2]{Corresponding author.}
\endgroup

\begin{abstract}
Large language model (LLM) capabilities have been associated with localized parameter regions, enabling targeted model editing and intervention. However, it remains unclear whether a region identified before model optimization remains an effective intervention target afterward. We evaluate the functional persistence and parameter overlap of coding capability regions across three instruction-tuned models under quantization, pruning, and LoRA-based adaptation. Although re-extracted regions exhibit substantial parameter reorganization, ablating the original region after optimization eliminates or nearly eliminates coding performance in almost all conditions, while magnitude-matched controls preserve performance. Partition analysis further shows that the shared subset of the original and re-extracted regions accounts for most of the observed ablation effect in the majority of conditions. These results demonstrate that parameter overlap and functional persistence are distinct: capability regions can remain causally effective intervention targets even when their parameter realization changes substantially.
\end{abstract}

\section{Introduction}

As large language models (LLMs) continue to improve across a wide range of tasks, understanding how their capabilities are internally organized has become an important research direction. Recent studies have shown that individual capabilities are localized to specific subsets of model parameters rather than being uniformly distributed throughout the model. This localization phenomenon has been observed for both factual knowledge \cite{dai2022knowledge} and task-specific capabilities \cite{christ2025math, kim2026exploring}, suggesting that diverse model capabilities are organized into identifiable capability regions.

The ability to identify capability regions has enabled a range of model intervention methods. Once identified, a capability region can be modified to edit factual knowledge, fine-tuned to acquire new skills while preserving existing capabilities, or removed to suppress undesirable behaviors. Although these methods serve different purposes, they all rely on the assumption that a capability region identified in one model remains valid when applied to the target model.

Despite this assumption, capability regions have been identified and validated only on the original model, without considering how models are transformed before deployment. In practice, deployed models rarely use their original weights. Instead, they are typically optimized for deployment, either to improve inference efficiency or to adapt the model to downstream tasks \cite{son2025analysis}. Deployment-time optimization reduces inference cost through techniques such as quantization and pruning, whereas training-time optimization adapts models through parameter-efficient methods such as Low-Rank Adaptation (LoRA). Although they serve different purposes, both modify the model parameters used during inference. Because capability regions are defined over model parameters, these weight modifications may alter the regions identified before optimization. Whether capability regions remain valid after optimization therefore remains unclear.

This raises a fundamental question: \textit{Are capability regions robust to the weight modifications introduced during model optimization?} If capability regions are not robust, intervention methods that manipulate previously identified capability regions may no longer be effective in practice.
Neither prior work on capability localization nor research on model optimization answers this question. The former identifies and validates capability regions only on the original model, while the latter evaluates optimized models without examining whether capability regions remain robust after optimization.

\begin{figure*}[t]
    \centering
    \includegraphics[width=0.8\linewidth]{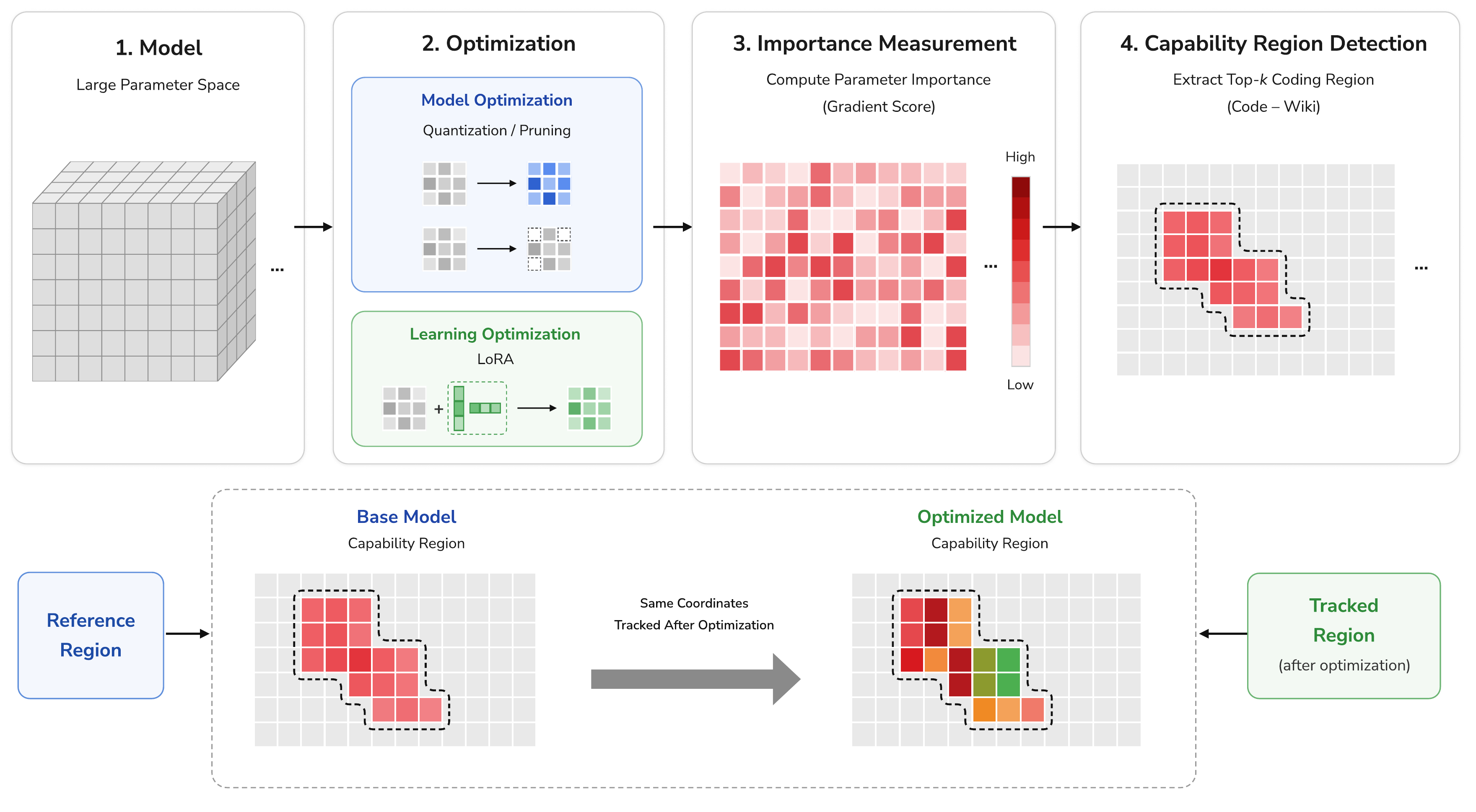}
    \caption{Overview of the proposed framework for evaluating capability region stability under model optimization. Capability regions are extracted using parameter importance scores and evaluated under quantization, pruning and LoRA.}
    \label{fig:fig1}
\end{figure*}

To answer this question, we consider robustness from two complementary perspectives. Functional persistence asks whether a capability region continues to support the same capability after optimization, whereas parameter overlap asks whether optimization preserves the parameters associated with that capability. Together, these perspectives capture both the functional and structural robustness of capability regions.

To study this question under realistic deployment conditions, we focus on coding capability. Among the capabilities for which localized parameter regions have been identified, coding models are routinely subjected to both deployment-time optimization for efficient serving and repeated parameter-efficient adaptation for specialized programming tasks. This makes coding a realistic and practically important setting for evaluating whether capability regions remain usable after optimization.

We make three main contributions. First, we introduce a framework that separates the functional persistence of a capability region from its parameter overlap after model optimization. Second, across three models and seven optimization or adaptation methods, we show that reference regions remain effective intervention targets despite substantial variation in their re-extracted coordinates. Third, by partitioning reference and optimized regions, we find that the observed causal effect is concentrated primarily in their shared subset rather than in newly selected coordinates.

\section{Related Work}
\label{sec:2}

\paragraph{Localizing Capabilities in Model Parameters}

Recent studies have shown that the capabilities of large language models are localized to identifiable subsets of model parameters rather than being uniformly distributed across the network \cite{geva2021transformer}. Initially demonstrated for factual knowledge \cite{meng2022locating, dai2022knowledge}, capability localization has since been extended to task-specific capabilities, including mathematical reasoning \cite{christ2025math}, coding \cite{kim2026exploring}, and multilingual competence \cite{tang2024language}. Across these studies, capability regions occupy only a small fraction of the model parameters and are validated by showing that removing the identified region selectively impairs the target capability while largely preserving general language ability.

Once identified, capability regions become targets for model intervention. Prior work has used them to edit factual knowledge \cite{meng2022mass}, suppress targeted behaviors \cite{pochinkov2024dissecting}, add capabilities through localized fine-tuning \cite{xu2025let}, and preserve safety-related behaviors \cite{wei2024assessing}. These interventions assume that a capability region identified in one model remains the correct target after the model has been modified. Prior work has examined this assumption under fixed model weights.

\paragraph{Optimizing Models for Deployment}

Models deployed in practice rarely use their original weights. Model optimization reduces inference cost through quantization and pruning \cite{han2015learning, frantar2022gptq, frantar2023sparsegpt, lin2024awq, sun2024simple}, and training optimization adapts models through parameter-efficient methods such as LoRA \cite{hu2021lora}. As a result, the parameters used at inference differ from those on which a capability region was originally identified.

A related line of work has analyzed the effects of model optimization on downstream performance \cite{lee2026faster}. Compression \cite{jaiswal2024compressing, marchisio2024does}, parameter-efficient adaptation, and weight-space interventions combined with compression \cite{kolbeinsson2025composable, zhang2025catastrophic} can all alter model capabilities. These studies, however, assess optimization through task performance rather than through the parameters responsible for a capability. Whether those parameters remain valid after optimization has therefore not yet been examined.

We study this question from two complementary perspectives. The first is \emph{functional persistence}, which asks whether a region identified in the original model still supports the capability after optimization. The second is \emph{region overlap}, which measures how much of the original region is recovered when capability region extraction is repeated on the optimized model.

\section{Method}
\label{sec:3}

We test whether a capability region identified in one model continues to support the same capability in the optimized model that is deployed in its place. Our methodology proceeds in three stages. We first remove the original capability region in the optimized model, then re-extract the capability region and compare it with the original, and finally partition the two regions into their shared and non-shared parameters and remove each separately.

\subsection{Extracting a Capability Region}
\label{sec:3.1}

The same capability region extraction procedure is used in every experiment. The importance score follows \cite{kim2026exploring}, and the region definition follows \cite{christ2025math}. Let $\theta \in \mathbb{R}^{P}$ denote the flattened parameter vector over all transformer-layer parameters, where $P$ is the number of such parameters, and let $\mathcal{D}$ denote a corpus partitioned into domains $d$. The extraction begins by estimating how important each parameter is for a given corpus. We compute an importance score $I_{\mathcal{D}}(\theta)$ for each parameter as

\begin{equation}
I_{\mathcal{D}}(\theta) \;=\; \sum_{d \in \mathcal{D}}
  \Bigl| \, \theta \odot \!\!\sum_{x \in d} \nabla_{\theta}\,\mathcal{L}(x) \Bigr| ,
\label{eq:importance}
\end{equation}

where $\mathcal{L}$ denotes the next-token prediction loss computed using causal language modeling, and $\odot$ denotes the elementwise product. 

A capability region is built from the parameters that are highly important for the target capability but not for general language. We first form a candidate region

\begin{equation}
R \;=\; \mathrm{top}_{k}\bigl(I_{\mathcal{C}}\bigr) \;\setminus\;
                \mathrm{top}_{k}\bigl(I_{\mathcal{W}}\bigr)  ,
\label{eq:region}
\end{equation}

where $I_{\mathcal C}$ and $I_{\mathcal W}$ are the importance scores computed from the code corpus $\mathcal C$ and the general-text corpus $\mathcal W$, respectively. The operator $\mathrm{top}_{k}$ selects the highest-scoring $kP$ parameters using a single global ranking rather than ranking within each tensor. The same fraction $k$ is used for both rankings.

\subsection{Reference and Optimized Models}
\label{sec:3.2}

For every optimization, we define a pair of models. We call the model from which a capability region is extracted the \emph{reference model} $M_{\mathrm{ref}}$, and its optimized counterpart the \emph{optimized model} $M_{\mathrm{opt}}$.

Under quantization and pruning, the weights of a trained model are modified without further training. Quantization lowers the numerical precision of the weights, whereas pruning sets a fraction of them to zero while preserving the model architecture. Here, $M_{\mathrm{ref}}$ is the released checkpoint and $M_{\mathrm{opt}}$ is its quantized or pruned counterpart.

Under LoRA adaptation, the training procedure is changed. Full fine-tuning updates every weight, whereas LoRA freezes the pretrained weights and learns a low-rank adaptation instead. Here, $M_{\mathrm{ref}}$ is the fully fine-tuned model and $M_{\mathrm{opt}}$ is a LoRA model trained from the same pretrained checkpoint using the same training data and objective.

These optimizations differ in nature. Model optimization modifies the weights of a trained model, whereas training optimization trains a separate model from the same pretrained checkpoint. Despite these differences, parameters retain a one-to-one correspondence between $M_{\mathrm{ref}}$ and $M_{\mathrm{opt}}$: quantization and pruning preserve the parameter layout, and LoRA adapters are merged into the base weights before capability region extraction. This correspondence allows the same parameters to be ablated in both models.

\subsection{Three Measurements}
\label{sec:3.3}

Ablating a set of parameters from a model means setting those parameters to zero and evaluating the result. Because optimization changes coding accuracy before any ablation is applied, we report both the absolute accuracy after ablation and its \emph{relative retention}, the fraction of the model's original accuracy that remains.

\paragraph{Functional persistence}
The first measurement asks whether a capability region identified in the reference model still supports the capability after optimization. $R_{\mathrm{ref}}$ is extracted once from $M_{\mathrm{ref}}$, transferred to $M_{\mathrm{opt}}$ through the parameter correspondence described in Section~\ref{sec:3.2}, and ablated there without re-extraction. If code generation collapses, the original region still supports the capability. Otherwise, the capability can no longer be removed by ablating that region alone.

\paragraph{region overlap}
Functional persistence does not reveal whether the same parameters are selected if the region is extracted again after optimization. We therefore repeat the extraction on $M_{\mathrm{opt}}$ to obtain $R_{\mathrm{opt}}$ and measure the fraction of the original region that is recovered,

\begin{equation}
O = \frac{|R_{\mathrm{ref}} \cap R_{\mathrm{opt}}|}{|R_{\mathrm{ref}}|},
\label{eq:overlap}
\end{equation}

which measures how much of the original capability region is recovered after optimization.

\paragraph{Parameter Partition}
When the two regions differ, the overlap alone does not identify which parameters still support the capability. We therefore partition them into the shared parameters
($S = R_{\mathrm{ref}} \cap R_{\mathrm{opt}}$), the disappeared parameters ($D = R_{\mathrm{ref}} \setminus R_{\mathrm{opt}}$), and the newly selected parameters ($N = R_{\mathrm{opt}} \setminus R_{\mathrm{ref}}$). Each subset is ablated separately in $M_{\mathrm{opt}}$ to determine whether the capability is supported by the shared parameters or by the parameters unique to either region.

\subsection{Controls and Validity}
\label{sec:3.4}

Removing $kP$ parameters changes a model regardless of whether those parameters belong to a capability region. Every ablation is therefore compared with a magnitude-matched control of the same size. The control matches the region in the number of parameters selected from each tensor and in the distribution of parameter magnitudes, while remaining disjoint from the capability region. It is used for all region-ablation experiments except the partition analysis in Section~\ref{sec:5.4}.

Model conditions that already fail at code generation are excluded. Otherwise, a failed ablation could simply reflect a failed model rather than the capability region being tested. An ablation is labelled as a global collapse if it severely degrades the model's general language capability. No condition reported in this paper meets this criterion.

\section{Experiment Setup}
\label{sec:4}

\subsection{Models and Optimizations}
\label{sec:4.1}

\paragraph{Models} 
We study three instruction-tuned models: Llama-3.1-8B-Instruct, Llama-3.2-3B-Instruct \cite{grattafiori2024llama}, and Mistral-7B-Instruct-v0.3\footnote{https://huggingface.co/mistralai/Mistral-7B-Instruct-v0.3}. The 3B model allows us to examine whether the observed effects depend on model scale rather than a single model size. We present the comprehensive specifications of all evaluated reference and optimized model pairs in Appendix~\ref{appendix:model-statistics}.

\paragraph{Quantization}
For quantization we use 4-bit weight-only round-to-nearest (RTN), GPTQ \cite{frantar2022gptq}, and AWQ \cite{lin2024awq}. 

\paragraph{Pruning}
For pruning we use unstructured magnitude pruning \cite{han2015learning}, Wanda \cite{sun2024simple}, and SparseGPT \cite{frantar2023sparsegpt}, each applied layer by layer at a sparsity of $0.3$.

\paragraph{Fine-tuning and LoRA}

We compare full fine-tuning and LoRA \cite{hu2021lora} by training both models from the same pretrained checkpoint on the same instruction-following dataset under the same objective. Both models are trained for three epochs on 500 general-purpose instruction-following examples from Alpaca-cleaned \footnote{https://huggingface.co/datasets/yahma/alpaca-cleaned}. The two methods differ in how the parameter updates are parameterized. The reference model is obtained by full fine-tuning, whereas the optimized model uses LoRA with rank $16$, $\alpha=32$, and adapters on all linear projections, merged into the base weights before capability region extraction.

\subsection{Data}
\label{sec:4.2}

\paragraph{Region extraction data}

Importance scores are computed using two corpora: a code corpus and a general-text corpus. The code corpus spans ten programming languages\footnote{https://huggingface.co/datasets/nampdn-ai/tiny-codes} excluding Python, whereas the general-text corpus consists of Wikipedia articles\footnote{https://huggingface.co/datasets/wikimedia/wikipedia} in six natural languages. Python is excluded to avoid overlap with HumanEval.

\paragraph{Calibration data}

Two of the optimizations require no calibration data: RTN rounds each group of weights independently, and magnitude pruning ranks weights by absolute value alone. The remaining four methods are calibrated on the same dataset, using 128 sequences from the English C4 corpus\footnote{https://huggingface.co/datasets/allenai/c4}; Wanda and SparseGPT further process these as 2048-token crops. Holding the calibration data fixed ensures that differences among AWQ, GPTQ, Wanda, and SparseGPT reflect the optimization algorithms rather than the data used for calibration. Because the calibration data consists only of general text, neither the code corpus used for region extraction nor any evaluation benchmark is used during optimization \cite{dodge2021documenting}.

\paragraph{Evaluation benchmarks}
 
We evaluate four complementary aspects of model behavior: code generation (HumanEval) \cite{chen2021evaluating}, commonsense reasoning (HellaSwag) \cite{zellers2019hellaswag}, mathematical reasoning (GSM8K) \cite{cobbe2021training}, and language modeling quality (WikiText-2 perplexity) \cite{merity2016pointer}. HumanEval is evaluated on all 164 problems using greedy decoding (pass@1).

\subsection{Region Construction and Controls}
\label{sec:4.3}

\paragraph{Region construction}

Capability regions are extracted as described in Section~\ref{sec:3}. Unless otherwise stated, experiments are performed with $k \in \{0.1\%, 0.25\%, 0.5\%, 1\%\}$.

\paragraph{Counterfactual controls}
To verify that the impact of capability-region ablation stems specifically from the chosen parameters rather than general model damage, we compare region ablation against a magnitude-matched control of equal size. For each region $R(k)$, the control matches its parameter count per tensor and its distribution of parameter magnitudes, while remaining strictly disjoint from the capability region itself. This counterfactual design isolates parameter identity as the key variable, ensuring that observed performance drops cannot be explained simply by removing a certain volume of weights or targeting large-magnitude parameters.

\begin{table*}[t]
\centering
\scriptsize
\setlength{\tabcolsep}{6pt}
\renewcommand{\arraystretch}{0.95}
\begin{tabular}{@{}l@{\hspace{6pt}}r@{\hspace{10pt}}cccc@{\hspace{10pt}}cccc@{\hspace{10pt}}cccc@{}}
\toprule
 & & \multicolumn{4}{c}{Llama-3.1-8B-Instruct} & \multicolumn{4}{c}{Mistral-7B-Instruct-v0.3} & \multicolumn{4}{c}{Llama-3.2-3B-Instruct} \\
\cmidrule(l{2pt}r{2pt}){3-6}\cmidrule(l{2pt}r{2pt}){7-10}\cmidrule(l{2pt}r{2pt}){11-14}
 & \multicolumn{1}{c}{$k$ (\%)} & \multicolumn{1}{c}{\shortstack{Human\\Eval}} & \multicolumn{1}{c}{GSM8K} & \multicolumn{1}{c}{\shortstack{Hella\\Swag}} & \multicolumn{1}{c}{PPL$\,\downarrow$} & \multicolumn{1}{c}{\shortstack{Human\\Eval}} & \multicolumn{1}{c}{GSM8K} & \multicolumn{1}{c}{\shortstack{Hella\\Swag}} & \multicolumn{1}{c}{PPL$\,\downarrow$} & \multicolumn{1}{c}{\shortstack{Human\\Eval}} & \multicolumn{1}{c}{GSM8K} & \multicolumn{1}{c}{\shortstack{Hella\\Swag}} & \multicolumn{1}{c}{PPL$\,\downarrow$} \\
\midrule
\hspace{0.5em}Base & \emph{none} & \emph{62.8} & \emph{81.5} & \emph{78.6} & \phantom{0}\emph{8.29} & \emph{38.4} & \emph{54.0} & \emph{82.4} & \emph{6.26} & \emph{48.8} & \emph{68.5} & \emph{69.2} & \emph{13.28} \\
\addlinespace[1pt]
\hspace{1em}Region & 0.1 & 28.7 & 40.5 & 71.8 & 15.48 & \phantom{0}9.8 & 40.5 & 80.6 & 7.42 & 34.8 & 62.0 & 69.2 & 19.68 \\
 & 0.25 & \phantom{0}3.7 & 47.5 & 71.0 & 13.78 & \phantom{0}1.8 & 38.0 & 81.2 & 7.79 & 12.2 & 59.5 & 68.0 & 19.16 \\
 & 0.5 & \phantom{0}\textcolor{red}{0.0} & 18.0 & 68.4 & 15.25 & \phantom{0}\textcolor{red}{0.0} & 32.0 & 80.8 & 7.84 & \phantom{0}\textcolor{red}{0.0} & 55.5 & 67.4 & 21.47 \\
 & 1 & \phantom{0}\textcolor{red}{0.0} & 13.5 & 65.6 & 15.95 & \phantom{0}\textcolor{red}{0.0} & 30.0 & 76.0 & 7.93 & \phantom{0}\textcolor{red}{0.0} & 49.0 & 63.8 & 25.08 \\
\addlinespace[1pt]
\hspace{1em}Control & 0.1 & 62.2 & 77.0 & 78.0 & \phantom{0}8.49 & 39.6 & 43.5 & 81.4 & 6.33 & 43.9 & 67.0 & 69.2 & 13.77 \\
 & 0.25 & 58.5 & 79.5 & 77.6 & \phantom{0}8.79 & 37.2 & 52.0 & 80.8 & 6.44 & 43.9 & 72.0 & 68.4 & 14.25 \\
 & 0.5 & 59.8 & 74.5 & 77.8 & \phantom{0}9.43 & 32.3 & 39.0 & 80.4 & 6.57 & 47.0 & 63.5 & 66.6 & 14.94 \\
 & 1 & 47.0 & 62.5 & 75.8 & 10.61 & 29.9 & 34.5 & 79.2 & 6.81 & 35.4 & 64.0 & 66.8 & 16.76 \\
\addlinespace[2pt]
\hspace{0.5em}Full FT & \emph{none} & \emph{47.6} & \emph{72.5} & \emph{78.6} & \phantom{0}\emph{8.36} & \emph{25.6} & \emph{22.0} & \emph{79.8} & \emph{6.71} & \emph{39.0} & \emph{66.5} & \emph{70.4} & \emph{12.14} \\
\addlinespace[1pt]
\hspace{1em}Region & 0.1 & 28.7 & 57.0 & 76.2 & 10.94 & \phantom{0}9.8 & 11.5 & 79.2 & 7.56 & 31.7 & 61.0 & 67.6 & 15.65 \\
 & 0.25 & \phantom{0}\textcolor{red}{0.0} & 24.0 & 74.8 & 13.41 & \phantom{0}7.9 & 14.0 & 75.6 & 7.78 & 29.9 & 60.5 & 69.2 & 15.46 \\
 & 0.5 & \phantom{0}\textcolor{red}{0.0} & 13.5 & 73.4 & 14.53 & \phantom{0}3.7 & 14.0 & 75.6 & 7.98 & \phantom{0}9.1 & 43.5 & 68.8 & 16.44 \\
 & 1 & \phantom{0}\textcolor{red}{0.0} & \phantom{0}4.5 & 69.0 & 15.32 & \phantom{0}0.6 & 11.5 & 74.0 & 8.31 & \phantom{0}\textcolor{red}{0.0} & 41.5 & 66.0 & 18.03 \\
\addlinespace[1pt]
\hspace{1em}Control & 0.1 & 51.2 & 68.0 & 78.0 & \phantom{0}8.57 & 22.0 & 22.0 & 78.8 & 6.78 & 32.9 & 60.5 & 70.0 & 12.37 \\
 & 0.25 & 50.6 & 68.5 & 78.4 & \phantom{0}8.89 & 19.5 & 24.5 & 78.6 & 6.83 & 32.3 & 65.5 & 69.2 & 12.61 \\
 & 0.5 & 42.7 & 63.0 & 77.2 & \phantom{0}9.43 & 17.7 & 17.0 & 78.2 & 7.00 & 29.9 & 60.5 & 69.0 & 13.08 \\
 & 1 & 31.1 & 52.0 & 76.8 & 10.54 & 16.5 & 17.5 & 77.2 & 7.24 & 21.3 & 55.5 & 67.2 & 13.93 \\
\bottomrule
\end{tabular}
\caption{\textbf{Ablating the capability region removes code generation, whereas a magnitude-matched control does not.} Both reference models are evaluated across all values of $k$. The capability region and the matched control remove the same number of parameters. Cells report absolute evaluation scores.}
\label{tab:tab1}
\end{table*}

\section{Results}
\label{sec:5}

Models are rarely deployed in the exact form where a capability region was originally identified. Before deployment, model weights typically undergo optimizations such as quantization, pruning, or LoRA-based adaptation. To assess whether a capability region remains valid after optimization, we first evaluate its functional persistence and then its parameter overlap. Functional persistence examines whether ablating the original region still eliminates the target capability, while parameter overlap measures how its structural location changes. We organize our analysis around four research questions (RQs):

\begin{itemize}
    \item \textbf{RQ1.} Does ablating the extracted capability region eliminate coding?
    \item \textbf{RQ2.} Does the capability region retain coding after optimization?
    \item \textbf{RQ3.} Is the capability region positionally preserved after optimization?
    \item \textbf{RQ4.} Which parameters carry the capability after optimization?
\end{itemize}

Together, these analyses determine whether capability regions remain usable for model intervention after model optimization and explain how optimization changes their parameter realization.

\begin{figure*}[t]
    \centering
    \includegraphics[width=\linewidth]{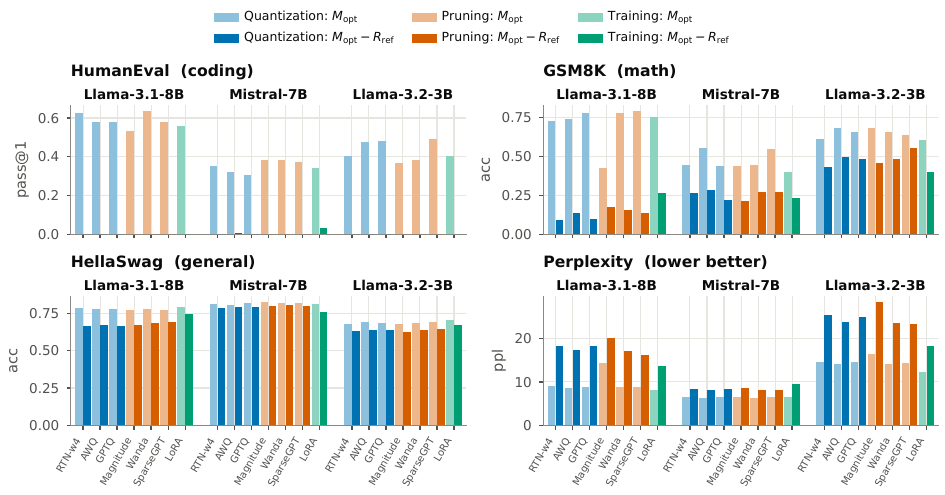}
    \caption{\textbf{Transfer results.} Each optimized model appears twice: $M_{\mathrm{opt}}$ (light) and $M_{\mathrm{opt}}-R_{\mathrm{ref}}$ (solid), evaluated at the reference model's $k^{\star}$. Colors indicate the optimization family. Ablating the transferred reference region substantially reduces HumanEval while largely preserving HellaSwag and perplexity.}
    \label{fig:fig2}
\end{figure*}

\subsection{RQ1: Does ablating the extracted capability region eliminate coding?}
\label{sec:5.1}

\paragraph{Reference models}
Throughout this subsection, $R(k)$ is extracted and ablated on the same model to evaluate the intrinsic effectiveness of the extraction rule itself. We analyze the base models and their fully fine-tuned counterparts as reference models ($M_{\mathrm{ref}}$), from which all capability regions in Sections~\ref{sec:5.2}--\ref{sec:5.4} are extracted. Table~\ref{tab:tab1} reports the ablation results across the evaluated range of $k$. On the base model, ablating the region produced by $k{=}0.5\%$ drives HumanEval to exactly zero across all three models. A similar trend is observed in the fully fine-tuned models, where region ablation severely impairs coding capability.

\paragraph{The effect is not a size or magnitude artifact}
As constructed in Section~\ref{sec:4.3}, the magnitude-matched control isolates the causal impact of the specific parameter set $R(k)$. Its substantially weaker effect rules out two alternative explanations: that ablating this many parameters is inherently sufficient to eliminate code generation, or that the performance drop is merely an artifact of removing large-magnitude weights. What eliminates code generation is therefore which parameters are ablated, not how many are ablated or how large they are.

\paragraph{Choosing the evaluation point}
Because the effect varies with $k$, we define the $k^{\star}$ for Sections~\ref{sec:5.2}--\ref{sec:5.4} as the smallest $k$ where HumanEval reaches zero while still passing the global-collapse check of Section~\ref{sec:3.4}. For conditions where HumanEval does not reach zero within the evaluated grid, we report results at the largest evaluated $k$ instead.

\paragraph{What survives}
Code performance drops sharply while the model remains a functioning language model. On the reference models, HellaSwag stays within a few points of its pre-ablation accuracy at $k^{\star}$ and never approaches chance anywhere in the grid, while WikiText-2 perplexity less than doubles, as shown in Table~\ref{tab:tab1}. Across all reference model conditions, no ablation triggers the global-collapse criterion of Section~\ref{sec:3.4}, confirming that the ablation selectively targets coding capability rather than causing general model failure. GSM8K, in contrast, declines in every condition, indicating that the extracted region is code-biased rather than code-exclusive.

Together, these analyses demonstrate that the extraction rule identifies parameter regions that causally support coding capability on reference models, establishing a validated baseline for subsequent analysis.

\subsection{RQ2: Does the capability region retain coding after optimization?}
\label{sec:5.2}

We transfer $R_{\mathrm{ref}}$ verbatim into $M_{\mathrm{opt}}$ through the parameter correspondence described in Section~\ref{sec:3.2} and ablate it without re-extraction. Because the coordinates remain fixed prior to model optimization, this procedure directly evaluates functional persistence. Figure~\ref{fig:fig2} summarizes the results across all $21$ optimized model pairs, with corresponding magnitude-matched controls provided in Appendix~\ref{appendix:transfer-controls}.

\paragraph{Quantization}
At the $k^{\star}$ of the reference model, transferring $R_{\mathrm{ref}}$ into quantized models consistently drives HumanEval to zero or near zero. RTN, AWQ, and GPTQ all display this same catastrophic drop in coding accuracy, while non-coding capabilities remain largely intact and matched controls retain more coding performance.

\paragraph{Pruning}
Pruning via magnitude, Wanda, and SparseGPT exhibits the same targeted drop in coding accuracy. Although pruning zeros out weights, overlapping zero-coordinates never exceed $3.5\%$ of the region in any cell, as detailed in Appendix~\ref{appendix:transfer-controls}, confirming that the collapse stems from targeted ablation rather than prior compression.

\paragraph{LoRA}
Carrying the reference region into LoRA-adapted models drives HumanEval to zero on both Llama models and to under a tenth of its pre-ablation score on Mistral-7B. Meanwhile, general language performance is preserved and magnitude-matched controls stay at pre-ablation levels, confirming that capability regions retain their functional necessity under rank-restricted adaptation.

\paragraph{Functional persistence and transition}
Across all $21$ deployment pairs, transferring $R_{\mathrm{ref}}$ reduces coding performance to zero or near zero while magnitude-matched controls retain more coding performance (Appendix~\ref{appendix:transfer-controls}), confirming that capability regions functionally retain their necessity after optimization as valid intervention targets. However, functional persistence does not imply region overlap, as the region may have shifted during optimization. Confirming whether the coordinates remain invariant or move requires re-extracting $R(k)$ directly on $M_{\mathrm{opt}}$, which we examine in Section~\ref{sec:5.3} and Section~\ref{sec:5.4}.

\begin{figure}[t]
    \centering
    \includegraphics[width=0.9\linewidth]{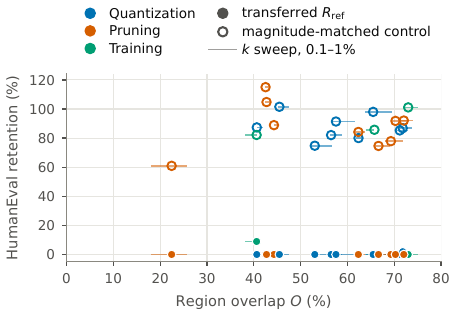}
    \caption{\textbf{Region overlap does not predict functional persistence.}
Each condition is shown twice, once for the transferred region and once for the magnitude-matched control, both evaluated at its own $k^{\star}$. Horizontal bars indicate the range of region overlap across the evaluated $k$ values.}
    \label{fig:fig3}
\end{figure}

\subsection{RQ3: Is the capability region positionally preserved after optimization?}
\label{sec:5.3}

Section~\ref{sec:5.2} leaves open whether the optimized model kept the region in place or moved it while the transferred coordinates remained sufficient. Separating them requires re-extracting $R_{\mathrm{opt}}$ on each $M_{\mathrm{opt}}$ to measure the region overlap $O$ as formulated in Eq.~\ref{eq:overlap}. To first verify that each re-extracted region $R_{\mathrm{opt}}$ effectively isolates coding capability, we confirm its ablation validity in Appendix~\ref{appendix:reextracted} before analyzing region overlap. The overlap is interpretable because ablating $R_{\mathrm{opt}}$ on its own model at $k^{\star}$ drives HumanEval to zero in $15$ of $21$ pairs, while leaving at most $3$ points in $5$ others, as documented across Table~\ref{tab:dose-quant} through Table~\ref{tab:dose-lora}. The sole exception is the Mistral-7B training pair with $9.1$ points, the same condition whose reference model never reaches zero anywhere on the grid. Furthermore, as shown in Appendix~\ref{appendix:seed-stability}, re-estimating importance on the same model yields a mean overlap of at least $92.7\%$. In contrast, the observed overlap at $k^\star$ has a median of about $60\%$ and ranges from $22.4\%$ to $73.0\%$, suggesting that sampling variability alone cannot account for the positional differences.

\paragraph{The overlap is a property of the pair, not of the threshold}
Sweeping $k$ over a factor of ten in Table~\ref{tab:recovery} moves overlap by a median of only $2.7$ points, full threshold-sweep statistics in Appendix~\ref{appendix:overlap}, while $R_{\mathrm{opt}}$ remains close to $R_{\mathrm{ref}}$. The approximately $50$-point spread across pairs confirms that how much of $R_{\mathrm{ref}}$ an optimized model reselects is a property of that model and optimization, not of thresholding choices.

\paragraph{The overlap does not predict function}
Figure~\ref{fig:fig3} plots region overlap against HumanEval retention at $k^{\star}$. Across the entire range, transferred regions sit at the floor with remaining near zero, while matched controls stay between $60.9\%$ and $115.0\%$. Region overlap and functional persistence are therefore different quantities, and the former is not a proxy for the latter. Section~\ref{sec:5.4} resolves which part of $R_{\mathrm{ref}}$ drives this persistent effect.

\begin{figure}[t]
    \centering
    \includegraphics[width=0.9\linewidth]{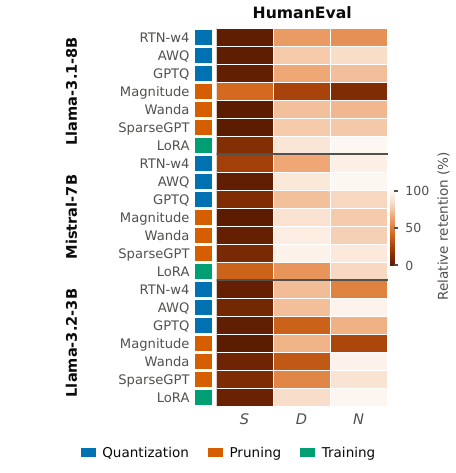} 
    \caption{\textbf{Shared parameters usually have the strongest ablation effect.} Each row shows separate ablations of shared $S$, disappeared $D$, and newly selected $N$ parameters in $M_{\mathrm{opt}}$. Darker cells indicate lower HumanEval retention. Ablating $S$ causes the largest decrease in $20$ of $21$ pairs.}
    \label{fig:fig4}
\end{figure}

\subsection{RQ4: Which parameters carry the capability after optimization?}
\label{sec:5.4}

\paragraph{Shared parameters usually have the strongest ablation effect}
We partition the union of $R_{\mathrm{ref}}$ and $R_{\mathrm{opt}}$ into the shared parameters $S$, disappeared parameters $D$, and newly selected parameters $N$, as defined in Section~\ref{sec:3.3}, and ablate each subset separately in $M_{\mathrm{opt}}$ at the reference model's $k^{\star}$. Full partition results are reported in Appendix~\ref{appendix:partition}. As shown in Table~\ref{tab:partition} and Figure~\ref{fig:fig4}, ablating $S$ alone drives HumanEval to zero in only $4$ of $21$ pairs. Nevertheless, it causes a larger decrease than ablating either $D$ or $N$ in $20$ pairs, while ablating $D$ or $N$ alone retains more than half of the pre-ablation score in most cases.

\paragraph{The strongest effect can occur in the smallest subset}
Because $S$ is the largest partition in $14$ pairs, its effect could be a size artifact. However, $S$ is the smallest partition in the seven remaining pairs, yet remains the most destructive in six of them. For instance, in Llama-3.2-3B under GPTQ, ablating $S$ drops HumanEval performance near zero, whereas ablating the substantially larger $D$ and $N$ partitions preserves considerable performance. These results show that ablating a larger subset does not necessarily cause a greater loss of coding capability.

\paragraph{An exception with stronger effects outside the shared subset}
The only exception is Llama-3.1-8B under magnitude pruning, where the two regions overlap the least. In this case, removing $S$ causes less damage than removing either $D$ or $N$, and no single subset reduces HumanEval to zero. This shows that the strongest ablation effect need not lie in the shared parameters.

\section{Conclusion}

We find that coding capability regions identified before optimization generally remain causally effective intervention targets after quantization, pruning, and LoRA-based adaptation. This functional persistence holds despite substantial variation in the parameters selected when the region is re-extracted, demonstrating that parameter overlap and intervention effectiveness capture distinct properties. Partition analysis further shows that ablating the shared subset of the original and re-extracted regions generally produces the largest decrease in coding performance among the three subsets. These results support evaluating capability localization not only through parameter similarity, but also through the transferability of interventions across model transformations.

\section*{Limitations}

Compute constraints restrict our evaluation to three 3B–8B English models across two families. Each compression method is evaluated at a single intensity, namely 4-bit quantization and 30\% unstructured pruning, which limits our ability to map region stability under stronger optimization. In addition, all cells represent single runs under a four-point $k$ grid, so we report no variance and treat $k^{\star}$ values as interval-censored rather than exact thresholds.

\section*{Ethical Statement}
We confirm that all models and datasets were utilized in strict accordance with their respective licenses, and our experimental evaluations were conducted objectively to ensure fairness and avoid bias.

\section*{Acknowledgments}
This work was supported by Korea Internet \& Security Agency(KISA) grant funded by the Korea government(PIPC) (No.RS-2026-25526342, Development of Technologies for Preventing Sensitive Information Inference and Risk Assessment in Foundation Model Operations). This work was also supported by the National Research Foundation of Korea(NRF) grant funded by the Korea government(MSIT) (RS-2026-25483747). In addition, this research was supported by Culture, Sports and Tourism R\&D Program through the Korea Creative Content Agency grant funded by the Ministry of Culture, Sports and Tourism in 2026 (Project Name: Develop AI agent technology to connect knowledge through public cultural facility-based discussion and communication, Project Number: RS-2026-25520645). Furthermore, this work was supported by the Institute of Information \& Communications Technology Planning \& Evaluation(IITP)-Innovative Human Resource Development for Local Intellectualization program grant funded by the Korea government(MSIT) (IITP-2026-RS-2022-00156360). Lastly, we gratefully acknowledge VESSL for providing GPU resources.


\bibliography{custom}

\clearpage
\appendix
\label{appendix}

\section{Additional Experimental Results}
\label{appendix:additional-results}

This appendix provides the complete results supporting the analyses in Section~\ref{sec:5}. We first report the model and optimization configurations, followed by the complete magnitude-matched transfer control results used in RQ2. We then present the condition-response results for regions re-extracted after each optimization along with the seed-resampling analysis of extraction stability. Finally, we provide the region-overlap statistics across all values of $k$ and the parameter-partition results used in RQ4.

\subsection{Model and Optimization Statistics}
\label{appendix:model-statistics}

\begin{figure*}[t]
    \centering
    \includegraphics[width=\linewidth]{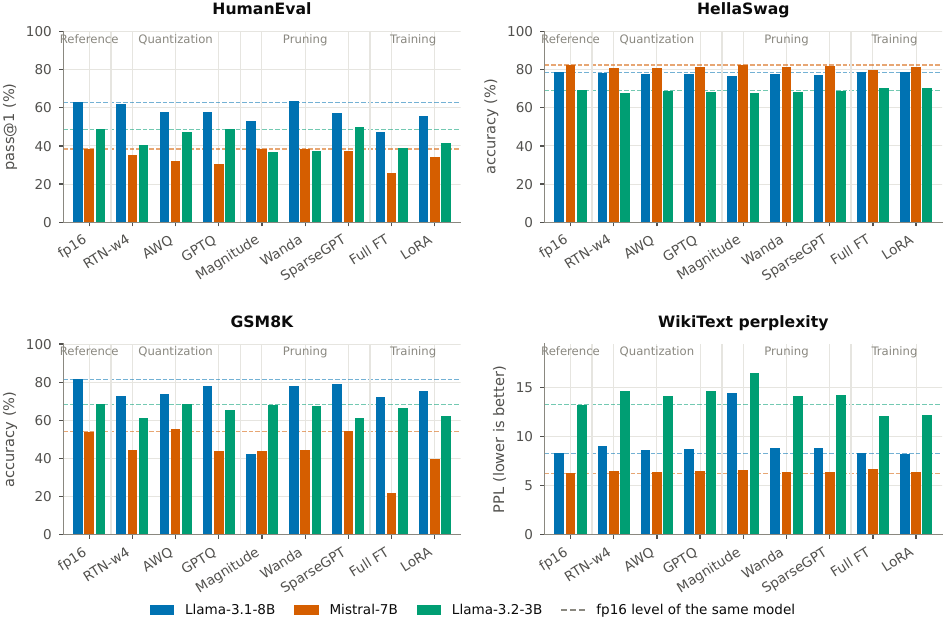}
    \caption{\textbf{Models and optimization configurations.} Summary of the reference models and their quantized, pruned, and LoRA-adapted counterparts evaluated in this study.}
    \label{fig:appendix1}
\end{figure*}

Figure~\ref{fig:appendix1} summarizes the model and optimization configurations used in our experiments. All optimized models retain a parameter-wise correspondence with their reference models, which allows reference regions to be transferred and ablated without re-extraction.

\subsection{Magnitude-Matched Transfer Controls}
\label{appendix:transfer-controls}

\begin{figure*}[t]
    \centering
    \includegraphics[width=\linewidth]{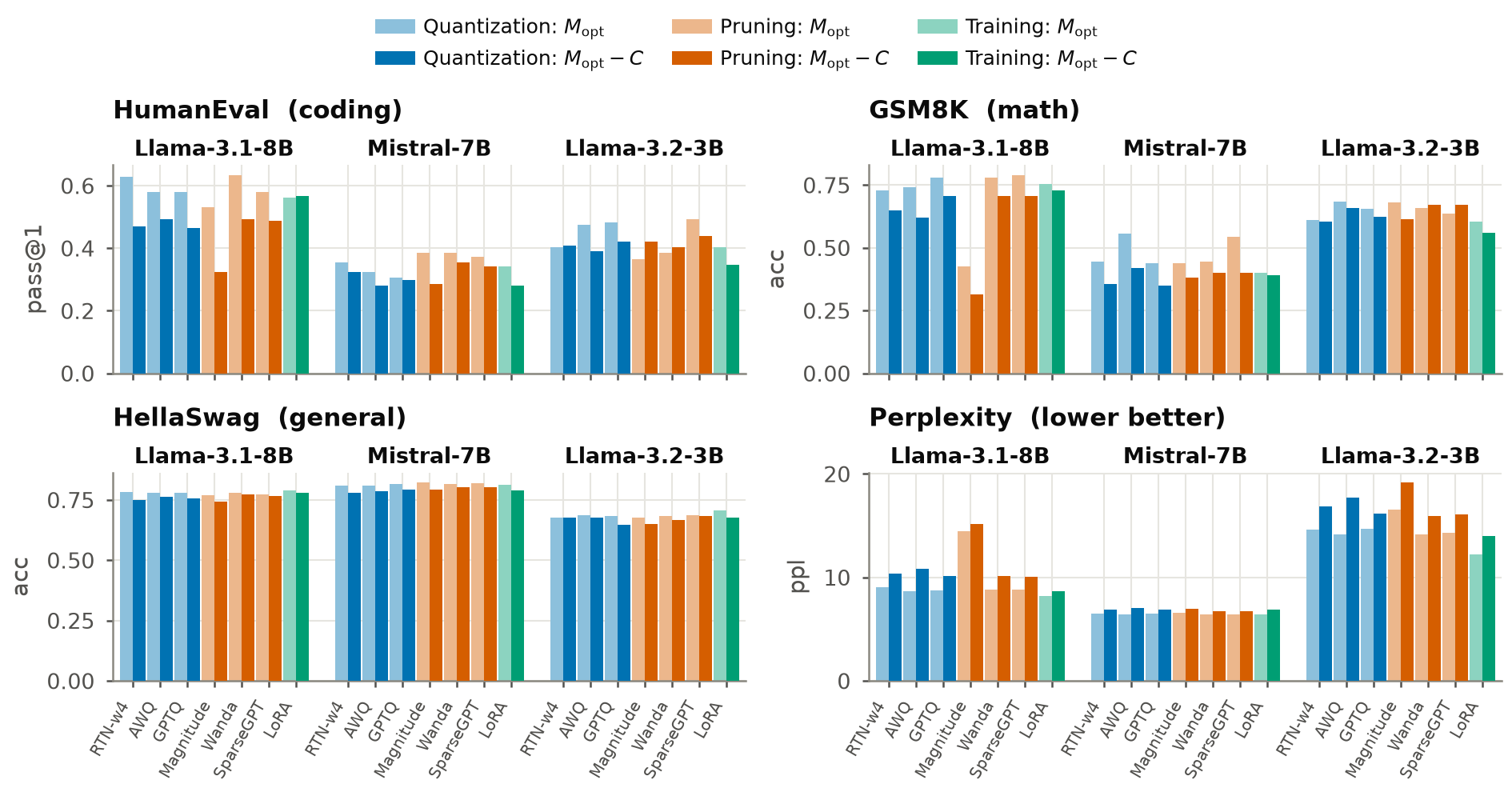}
    \caption{\textbf{Magnitude-matched controls for region transfer.}
Performance before ablation is compared with
performance after ablating a disjoint control set matched in parameter count,
tensor-wise allocation, and magnitude distribution.}
    \label{fig:magnitude_control}
\end{figure*}

Figure~\ref{fig:magnitude_control} provides the complete magnitude-matched control results for the transfer experiments in Section~\ref{sec:5.2}. Unlike ablating the transferred reference region, removing an equally sized set of parameters with a matched magnitude distribution generally preserves HumanEval. This confirms that the transferred-region effect cannot be explained by deletion size or parameter magnitude alone.

To assess the overlap with prior pruning, Table~\ref{tab:already-zero} reports the fraction of coordinates in $R_{\mathrm{ref}}$ already zeroed out before ablation. Across all reported pruning
conditions, this pre-existing sparsity remains negligible
(never exceeding $3.42\%$). This confirms that the intervention
newly zeros out the vast majority of the selected coordinates.

\begin{table}[t]
  \centering
  \small
  \setlength{\tabcolsep}{4pt}
  \begin{tabular}{@{}lrr@{}}
  \toprule
   & \multicolumn{2}{c}{Already zero (\% of $R_{\mathrm{ref}}$)} \\
  \cmidrule(l{2pt}r{2pt}){2-3}
   & $k{=}0.25\%$ & $k{=}0.5\%$ \\
  \midrule
  \multicolumn{3}{@{}l}{\textbf{Llama-3.1-8B}} \\
  \hspace{0.5em}Magnitude & 3.42 & \textbf{3.17} \\
  \hspace{0.5em}Wanda & 0.46 & \textbf{0.49} \\
  \hspace{0.5em}SparseGPT & 1.52 & \textbf{1.47} \\
  \multicolumn{3}{@{}l}{\textbf{Mistral-7B}} \\
  \hspace{0.5em}Magnitude & 1.65 & \textbf{1.60} \\
  \hspace{0.5em}Wanda & 0.91 & \textbf{0.97} \\
  \hspace{0.5em}SparseGPT & 1.23 & \textbf{1.29} \\
  \multicolumn{3}{@{}l}{\textbf{Llama-3.2-3B}} \\
  \hspace{0.5em}Magnitude & 1.41 & \textbf{1.47} \\
  \hspace{0.5em}Wanda & 0.44 & \textbf{0.49} \\
  \hspace{0.5em}SparseGPT & 0.97 & \textbf{1.04} \\
  \bottomrule
  \end{tabular}
  \caption{\textbf{Pre-existing zero coordinates in $R_{\mathrm{ref}}$ under pruning.} Fraction of transferred reference region coordinates that are already zero in $M_{\mathrm{opt}}$ prior to ablation. Across all conditions, this overlap remains under $3.42\%$, confirming that functional collapse is not an artifact of prior weight removal. Bold indicates $k^{\star}{=}0.5\%$.}
  \label{tab:already-zero}
  \end{table}

\subsection{Ablation of Re-extracted Regions}
\label{appendix:reextracted}

Tables~\ref{tab:dose-quant}--\ref{tab:dose-lora} report the complete
condition-response results obtained by extracting a new capability region directly
from each optimized model and ablating it on that same model. These experiments
verify that $R_{\mathrm{opt}}$ is itself a valid capability region before it is
compared with $R_{\mathrm{ref}}$. In nearly all conditions, increasing $k$
progressively reduces HumanEval to zero or near zero, while HellaSwag remains
substantially more stable. GSM8K also decreases in several conditions,
consistent with the code-biased rather than code-exclusive nature of the
extracted regions.

\paragraph{Quantization.}
Table~\ref{tab:dose-quant} presents the full results for RTN, AWQ, and GPTQ.
Despite the weight discretization introduced by each method, ablating the
region re-extracted from the quantized model consistently produces a sharp
reduction in coding performance.

\begin{table*}[t]
\centering
\scriptsize
\setlength{\tabcolsep}{6pt}
\renewcommand{\arraystretch}{0.95}
\begin{tabular}{@{}l@{\hspace{4pt}}r@{\hspace{9pt}}r@{\hspace{9pt}}r@{\hspace{9pt}}r@{\hspace{9pt}}r@{}}
\toprule
 & \multicolumn{1}{c}{$k$ (\%)} & \multicolumn{1}{c}{\shortstack{Human\\Eval}} & \multicolumn{1}{c}{\shortstack{Hella\\Swag}} & \multicolumn{1}{c}{GSM8K} & \multicolumn{1}{c}{PPL$\,\downarrow$} \\
\midrule
\multicolumn{6}{@{}l}{\textbf{Llama-3.1-8B-Instruct}} \\
\hspace{0.5em}RTN-W4 & 0.1 & 16.5 & 71.2 & 31.5 & 14.42 \\
 & 0.25 & 0.6 & 68.0 & 13.5 & 14.73 \\
 & 0.5 & \textcolor{red}{0.0} & 65.2 & 6.5 & 17.55 \\
 & 1 & \textcolor{red}{0.0} & 62.6 & 2.0 & 20.68 \\
\hspace{0.5em}AWQ & 0.1 & 13.4 & 70.2 & 48.0 & 13.08 \\
 & 0.25 & 0.6 & 68.2 & 25.0 & 14.65 \\
 & 0.5 & \textcolor{red}{0.0} & 64.8 & 15.5 & 15.86 \\
 & 1 & \textcolor{red}{0.0} & 63.2 & 3.5 & 18.29 \\
\hspace{0.5em}GPTQ & 0.1 & 9.1 & 73.4 & 47.5 & 12.34 \\
 & 0.25 & 0.6 & 68.2 & 28.0 & 13.89 \\
 & 0.5 & \textcolor{red}{0.0} & 68.2 & 15.0 & 15.32 \\
 & 1 & \textcolor{red}{0.0} & 66.8 & 9.0 & 15.94 \\
\addlinespace[2pt]
\multicolumn{6}{@{}l}{\textbf{Mistral-7B-Instruct-v0.3}} \\
\hspace{0.5em}RTN-W4 & 0.1 & 17.7 & 79.6 & 41.0 & 7.81 \\
 & 0.25 & 9.8 & 79.0 & 32.0 & 8.06 \\
 & 0.5 & 0.6 & 77.6 & 23.5 & 8.34 \\
 & 1 & \textcolor{red}{0.0} & 76.4 & 17.0 & 8.53 \\
\hspace{0.5em}AWQ & 0.1 & 9.8 & 79.4 & 38.0 & 7.64 \\
 & 0.25 & 5.5 & 79.8 & 35.5 & 7.93 \\
 & 0.5 & \textcolor{red}{0.0} & 78.4 & 31.0 & 8.01 \\
 & 1 & \textcolor{red}{0.0} & 76.6 & 23.5 & 8.27 \\
\hspace{0.5em}GPTQ & 0.1 & 7.3 & 79.0 & 22.5 & 8.26 \\
 & 0.25 & 1.8 & 78.6 & 29.5 & 8.65 \\
 & 0.5 & \textcolor{red}{0.0} & 78.8 & 19.0 & 8.78 \\
 & 1 & \textcolor{red}{0.0} & 75.6 & 17.0 & 8.84 \\
\addlinespace[2pt]
\multicolumn{6}{@{}l}{\textbf{Llama-3.2-3B-Instruct}} \\
\hspace{0.5em}RTN-W4 & 0.1 & 9.1 & 59.6 & 40.5 & 42.22 \\
 & 0.25 & \textcolor{red}{0.0} & 62.6 & 35.0 & 35.93 \\
 & 0.5 & \textcolor{red}{0.0} & 61.8 & 24.0 & 38.07 \\
 & 1 & \textcolor{red}{0.0} & 60.2 & 13.5 & 47.85 \\
\hspace{0.5em}AWQ & 0.1 & 36.6 & 66.6 & 61.5 & 20.98 \\
 & 0.25 & 12.2 & 66.2 & 55.0 & 21.33 \\
 & 0.5 & 1.2 & 64.2 & 52.5 & 24.10 \\
 & 1 & \textcolor{red}{0.0} & 61.4 & 40.0 & 28.30 \\
\hspace{0.5em}GPTQ & 0.1 & 28.7 & 66.2 & 60.0 & 20.57 \\
 & 0.25 & 5.5 & 64.6 & 60.0 & 21.47 \\
 & 0.5 & \textcolor{red}{0.0} & 66.4 & 49.0 & 22.64 \\
 & 1 & \textcolor{red}{0.0} & 60.8 & 43.0 & 26.80 \\
\bottomrule
\end{tabular}
\caption{\textbf{Effect of ablating re-extracted regions under 4-bit weight-only quantization.} Across model substrates, ablating the locally re-extracted capability region selectively reduces coding performance while HellaSwag remains comparatively more stable.}
\label{tab:dose-quant}
\end{table*}

\paragraph{Pruning.}
Table~\ref{tab:dose-prune} reports the corresponding results for magnitude
pruning, Wanda, and SparseGPT at 30\% unstructured sparsity. Re-extracted
regions remain strongly associated with coding performance across all three
pruning methods.

\begin{table*}[t]
\centering
\scriptsize
\setlength{\tabcolsep}{6pt}
\renewcommand{\arraystretch}{0.95}
\begin{tabular}{@{}l@{\hspace{4pt}}r@{\hspace{9pt}}r@{\hspace{9pt}}r@{\hspace{9pt}}r@{\hspace{9pt}}r@{}}
\toprule
 & \multicolumn{1}{c}{$k$ (\%)} & \multicolumn{1}{c}{\shortstack{Human\\Eval}} & \multicolumn{1}{c}{\shortstack{Hella\\Swag}} & \multicolumn{1}{c}{GSM8K} & \multicolumn{1}{c}{PPL$\,\downarrow$} \\
\midrule
\multicolumn{6}{@{}l}{\textbf{Llama-3.1-8B-Instruct}} \\
\hspace{0.5em}Magnitude & 0.1 & 6.7 & 72.2 & 4.0 & 21.34 \\
 & 0.25 & 1.2 & 72.2 & 4.0 & 24.09 \\
 & 0.5 & 1.2 & 68.0 & 3.5 & 23.11 \\
 & 1 & \textcolor{red}{0.0} & 69.0 & 3.0 & 22.46 \\
\hspace{0.5em}Wanda & 0.1 & 16.5 & 72.2 & 59.0 & 14.24 \\
 & 0.25 & \textcolor{red}{0.0} & 70.8 & 46.0 & 14.06 \\
 & 0.5 & \textcolor{red}{0.0} & 68.0 & 8.0 & 16.46 \\
 & 1 & \textcolor{red}{0.0} & 65.4 & 5.0 & 18.88 \\
\hspace{0.5em}SparseGPT & 0.1 & 14.6 & 71.6 & 49.0 & 12.44 \\
 & 0.25 & \textcolor{red}{0.0} & 71.4 & 34.0 & 13.70 \\
 & 0.5 & \textcolor{red}{0.0} & 68.2 & 18.0 & 14.77 \\
 & 1 & \textcolor{red}{0.0} & 65.2 & 9.5 & 15.77 \\
\addlinespace[2pt]
\multicolumn{6}{@{}l}{\textbf{Mistral-7B-Instruct-v0.3}} \\
\hspace{0.5em}Magnitude & 0.1 & 4.9 & 81.2 & 36.5 & 7.81 \\
 & 0.25 & 0.6 & 80.4 & 32.0 & 8.12 \\
 & 0.5 & \textcolor{red}{0.0} & 79.0 & 22.5 & 8.54 \\
 & 1 & \textcolor{red}{0.0} & 76.0 & 13.5 & 8.81 \\
\hspace{0.5em}Wanda & 0.1 & 16.5 & 81.0 & 37.0 & 7.35 \\
 & 0.25 & 6.1 & 79.8 & 36.0 & 7.71 \\
 & 0.5 & \textcolor{red}{0.0} & 79.0 & 32.0 & 7.89 \\
 & 1 & \textcolor{red}{0.0} & 75.2 & 23.5 & 8.37 \\
\hspace{0.5em}SparseGPT & 0.1 & 10.4 & 81.6 & 36.5 & 7.44 \\
 & 0.25 & 4.9 & 80.2 & 44.5 & 7.68 \\
 & 0.5 & \textcolor{red}{0.0} & 80.4 & 30.5 & 7.85 \\
 & 1 & \textcolor{red}{0.0} & 76.6 & 23.5 & 8.22 \\
\addlinespace[2pt]
\multicolumn{6}{@{}l}{\textbf{Llama-3.2-3B-Instruct}} \\
\hspace{0.5em}Magnitude & 0.1 & 1.2 & 59.4 & 15.0 & 50.19 \\
 & 0.25 & \textcolor{red}{0.0} & 61.2 & 36.0 & 42.65 \\
 & 0.5 & \textcolor{red}{0.0} & 58.0 & 19.5 & 43.60 \\
 & 1 & \textcolor{red}{0.0} & 54.4 & 2.0 & 50.50 \\
\hspace{0.5em}Wanda & 0.1 & 21.3 & 67.0 & 65.5 & 20.22 \\
 & 0.25 & 18.9 & 66.2 & 60.0 & 20.68 \\
 & 0.5 & 1.2 & 66.6 & 49.0 & 22.88 \\
 & 1 & \textcolor{red}{0.0} & 61.8 & 41.0 & 26.27 \\
\hspace{0.5em}SparseGPT & 0.1 & 26.2 & 66.4 & 61.0 & 20.71 \\
 & 0.25 & 5.5 & 66.6 & 57.0 & 21.02 \\
 & 0.5 & \textcolor{red}{0.0} & 65.4 & 51.0 & 22.36 \\
 & 1 & \textcolor{red}{0.0} & 61.4 & 45.0 & 25.94 \\
\bottomrule
\end{tabular}
\caption{\textbf{Effect of ablating re-extracted regions under unstructured pruning.} Across pruned substrates, zeroing out the locally re-extracted capability region selectively reduces coding performance while HellaSwag remains comparatively more stable.}
\label{tab:dose-prune}
\end{table*}

\paragraph{LoRA adaptation.}
Table~\ref{tab:dose-lora} reports the results for the LoRA-adapted models. The re-extracted regions eliminate coding on both Llama models at the evaluated grid endpoint and substantially reduce it on Mistral-7B, whose reference full-fine-tuned model also does not reach zero within the evaluated values of $k$.

\begin{table*}[t]
\centering
\scriptsize
\setlength{\tabcolsep}{6pt}
\renewcommand{\arraystretch}{0.95}
\begin{tabular}{@{}l@{\hspace{4pt}}r@{\hspace{9pt}}r@{\hspace{9pt}}r@{\hspace{9pt}}r@{\hspace{9pt}}r@{}}
\toprule
 & \multicolumn{1}{c}{$k$ (\%)} & \multicolumn{1}{c}{\shortstack{Human\\Eval}} & \multicolumn{1}{c}{\shortstack{Hella\\Swag}} & \multicolumn{1}{c}{GSM8K} & \multicolumn{1}{c}{PPL$\,\downarrow$} \\
\midrule
\multicolumn{6}{@{}l}{\textbf{Llama-3.1-8B-Instruct}} \\
\hspace{0.5em}LoRA & 0.1 & 34.8 & 75.0 & 36.0 & 11.76 \\
 & 0.25 & 3.0 & 72.8 & 37.5 & 12.40 \\
 & 0.5 & \textcolor{red}{0.0} & 70.2 & 10.0 & 14.25 \\
 & 1 & \textcolor{red}{0.0} & 66.8 & 9.5 & 16.02 \\
\addlinespace[2pt]
\multicolumn{6}{@{}l}{\textbf{Mistral-7B-Instruct-v0.3}} \\
\hspace{0.5em}LoRA & 0.1 & 30.5 & 80.0 & 37.5 & 6.92 \\
 & 0.25 & 27.4 & 79.2 & 35.5 & 7.08 \\
 & 0.5 & 15.9 & 79.0 & 34.5 & 7.37 \\
 & 1 & 9.1 & 78.2 & 32.5 & 7.67 \\
\addlinespace[2pt]
\multicolumn{6}{@{}l}{\textbf{Llama-3.2-3B-Instruct}} \\
\hspace{0.5em}LoRA & 0.1 & 34.8 & 68.8 & 63.5 & 15.45 \\
 & 0.25 & 28.0 & 69.6 & 55.0 & 15.92 \\
 & 0.5 & 9.8 & 68.4 & 47.0 & 17.09 \\
 & 1 & \textcolor{red}{0.0} & 67.4 & 41.5 & 18.47 \\
\bottomrule
\end{tabular}
\caption{\textbf{Effect of ablating re-extracted regions under LoRA fine-tuning.} Across LoRA-adapted substrates, zeroing out the locally re-extracted capability region reduces coding performance while HellaSwag remains comparatively more stable.}
\label{tab:dose-lora}
\end{table*}

\subsection{Stability of Region Extraction}
\label{appendix:seed-stability}

Table~\ref{tab:seed-floor} reports mean overlap above $92\%$
across all models and $k$ values when only importance-estimation samples change. This suggests that sampling variability alone cannot explain the much lower overlap after optimization.

\begin{table*}[t]
\centering
\small
\setlength{\tabcolsep}{4pt}
\begin{tabular}{@{}l@{\hspace{6pt}}rrrr@{}}
\toprule
 & \multicolumn{4}{c}{Seed-resampling overlap (\%)} \\
\cmidrule(l{2pt}r{2pt}){2-5}
 & $k{=}0.1\%$ & $0.25\%$ & $0.5\%$ & $1\%$ \\
\midrule
Llama-3.1-8B & 94.7 & 94.2 & \textbf{93.6} & 92.7 \\
Mistral-7B & 96.9 & 96.5 & \textbf{96.2} & 95.8 \\
Llama-3.2-3B & 96.3 & 95.9 & \textbf{95.3} & 94.6 \\
\bottomrule
\end{tabular}
\caption{\textbf{Re-extracting the region on an unchanged model reproduces almost all of it.} Each cell averages overlap across the three pairs formed by three seeds. The mean overlap deficit from re-extraction is at most $7.3$ percentage points, versus $27.0$--$77.6$ after optimization at $k^{\star}$ (Table~\ref{tab:recovery}).}
\label{tab:seed-floor}
\end{table*}

\subsection{Region Overlap across Thresholds}
\label{appendix:overlap}

Table~\ref{tab:recovery} reports region overlap and relative region size over the complete $k$ grid. Although overlap varies substantially across model--optimization pairs, it changes comparatively little as $k$ varies within a pair. The optimized and reference regions also remain similar in size. Thus, the observed structural differences primarily reflect changes in which parameters are selected rather than large changes in region cardinality.

\begin{table*}[t]
\centering
\small
\setlength{\tabcolsep}{5pt}
\begin{tabular}{@{}l@{\hspace{10pt}}rrrr@{\hspace{14pt}}rrrr@{}}
\toprule
 & \multicolumn{4}{c}{region overlap $O$ (\%)} & \multicolumn{4}{c}{Size ratio $|R_{\mathrm{opt}}|/|R_{\mathrm{ref}}|$} \\
\cmidrule(l{2pt}r{2pt}){2-5}\cmidrule(l{2pt}r{2pt}){6-9}
 & $k{=}0.1\%$ & $k{=}0.25\%$ & $k{=}0.5\%$ & $k{=}1\%$ & $0.1\%$ & $0.25\%$ & $0.5\%$ & $1\%$ \\
\midrule
\multicolumn{9}{@{}l}{\textbf{Llama-3.1-8B-Instruct}} \\
\hspace{0.5em}RTN-w4 & 56.7 & 54.3 & \textbf{53.0} & 52.1 & 1.030 & 1.013 & \textbf{0.998} & 0.980 \\
\hspace{0.5em}AWQ & 73.0 & 71.9 & \textbf{71.2} & 70.3 & 0.976 & 0.978 & \textbf{0.983} & 0.988 \\
\hspace{0.5em}GPTQ & 63.4 & 62.8 & \textbf{62.4} & 61.9 & 0.964 & 0.969 & \textbf{0.974} & 0.977 \\
\hspace{0.5em}Magnitude & 18.0 & 19.8 & \textbf{22.4} & 25.7 & 0.825 & 0.831 & \textbf{0.853} & 0.885 \\
\hspace{0.5em}Wanda & 71.9 & 70.4 & \textbf{69.3} & 68.0 & 0.974 & 0.977 & \textbf{0.984} & 0.987 \\
\hspace{0.5em}SparseGPT & 63.8 & 62.9 & \textbf{62.3} & 61.7 & 0.940 & 0.948 & \textbf{0.958} & 0.968 \\
\hspace{0.5em}LoRA & 72.0 & \textbf{73.0} & 74.0 & 75.0 & 1.014 & \textbf{1.012} & 1.010 & 1.010 \\
\multicolumn{9}{@{}l}{\textbf{Mistral-7B-Instruct-v0.3}} \\
\hspace{0.5em}RTN-w4 & 61.6 & 59.0 & \textbf{57.5} & 56.6 & 1.001 & 0.997 & \textbf{0.990} & 0.980 \\
\hspace{0.5em}AWQ & 73.9 & 72.7 & \textbf{71.8} & 70.8 & 0.975 & 0.980 & \textbf{0.981} & 0.983 \\
\hspace{0.5em}GPTQ & 69.5 & 67.2 & \textbf{65.5} & 63.8 & 1.070 & 1.041 & \textbf{1.025} & 1.013 \\
\hspace{0.5em}Magnitude & 69.1 & 67.8 & \textbf{66.6} & 65.7 & 0.991 & 0.995 & \textbf{0.994} & 0.995 \\
\hspace{0.5em}Wanda & 74.1 & 73.0 & \textbf{72.0} & 71.0 & 0.965 & 0.970 & \textbf{0.971} & 0.974 \\
\hspace{0.5em}SparseGPT & 72.6 & 71.3 & \textbf{70.2} & 69.2 & 0.968 & 0.973 & \textbf{0.977} & 0.982 \\
\hspace{0.5em}LoRA & 38.2 & 38.9 & 39.6 & \textbf{40.6} & 0.895 & 0.912 & 0.920 & \textbf{0.923} \\
\multicolumn{9}{@{}l}{\textbf{Llama-3.2-3B-Instruct}} \\
\hspace{0.5em}RTN-w4 & 47.6 & 46.2 & \textbf{45.5} & 45.1 & 1.090 & 1.091 & \textbf{1.083} & 1.073 \\
\hspace{0.5em}AWQ & 58.8 & 57.3 & \textbf{56.5} & 56.3 & 1.003 & 1.004 & \textbf{1.002} & 1.001 \\
\hspace{0.5em}GPTQ & 42.0 & 40.8 & \textbf{40.6} & 41.1 & 0.967 & 0.970 & \textbf{0.972} & 0.973 \\
\hspace{0.5em}Magnitude & 43.5 & 42.8 & \textbf{42.5} & 42.4 & 1.064 & 1.077 & \textbf{1.085} & 1.086 \\
\hspace{0.5em}Wanda & 42.7 & 42.4 & \textbf{42.7} & 43.6 & 0.968 & 0.965 & \textbf{0.963} & 0.962 \\
\hspace{0.5em}SparseGPT & 45.1 & 44.3 & \textbf{44.3} & 44.8 & 0.993 & 0.990 & \textbf{0.986} & 0.983 \\
\hspace{0.5em}LoRA & 64.1 & 63.9 & 64.5 & \textbf{65.8} & 0.955 & 0.952 & 0.955 & \textbf{0.957} \\
\bottomrule
\end{tabular}
\caption{\textbf{Region overlap and relative size after optimization across model pairs.} While sweeping the region size parameter $k$ barely shifts region overlap, optimization alters parameter selection far more than overall region size.}
\label{tab:recovery}
\end{table*}

\begin{figure*}[t]
    \centering
    \includegraphics[width=\linewidth]{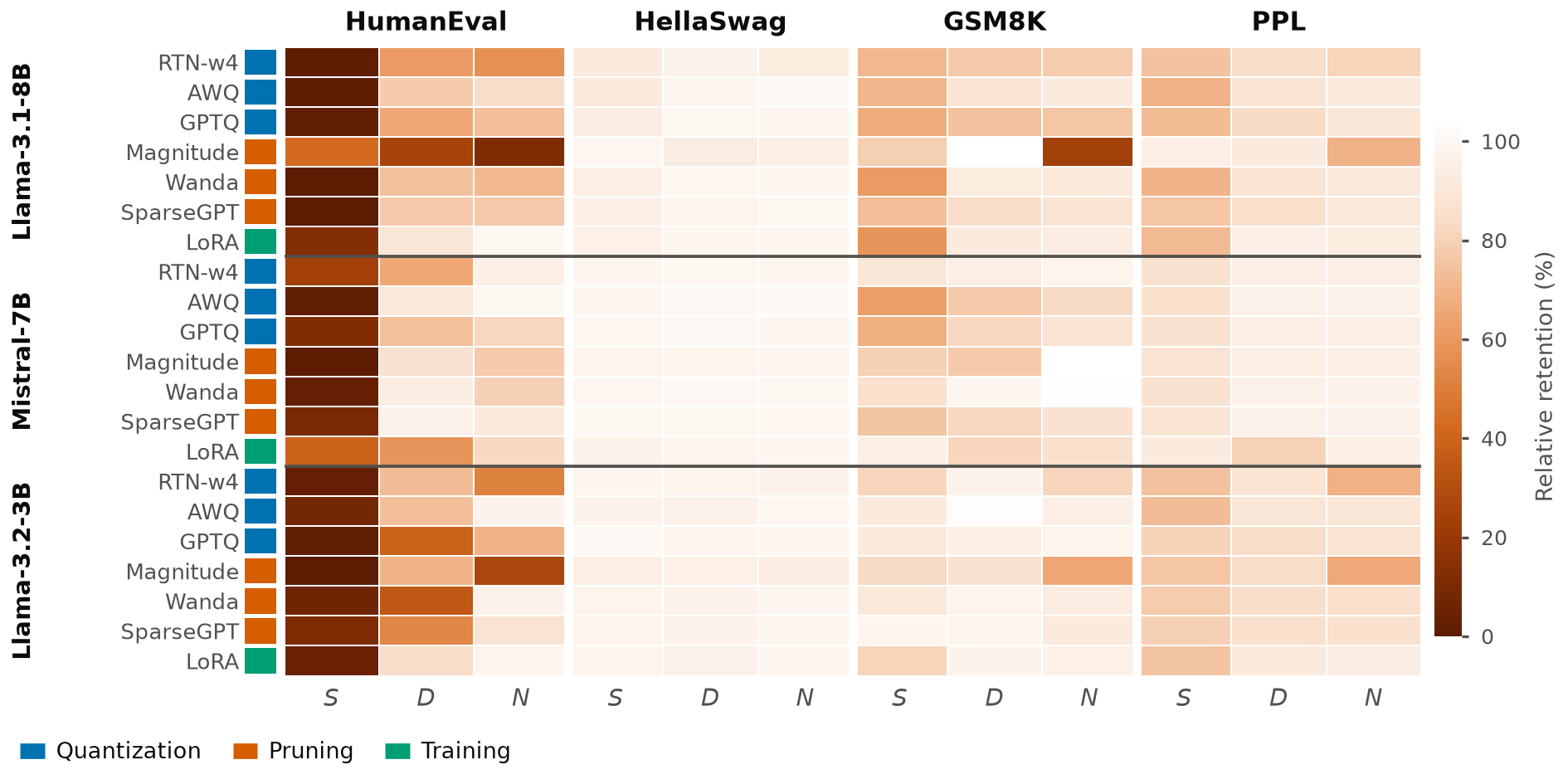}
    \caption{\textbf{Parameter-partition ablation results.}
Results for ablating the shared, disappeared, and newly selected parameter subsets across all model--optimization pairs.}
    \label{fig:S/N/R}
\end{figure*}

Figure~\ref{fig:S/N/R} additionally visualizes partition results. It shows that the strong functional persistence of $R_{\mathrm{ref}}$ is generally associated with the shared subset, whereas the disappeared and newly selected subsets individually produce weaker and more variable effects.

\subsection{Parameter-Partition Results}
\label{appendix:partition}

Table~\ref{tab:partition} reports the complete results obtained by separately ablating the shared parameters $S$, disappeared parameters $D$, and newly selected parameters $N$. These values underlie the analysis in Section~\ref{sec:5.4}. Across most model--optimization pairs, ablating $S$ causes the largest reduction in HumanEval, even when $S$ is smaller than one or both non-shared partitions.

\begin{table*}[t]
\centering
\scriptsize
\setlength{\tabcolsep}{3pt}
\renewcommand{\arraystretch}{0.9}
\begin{tabular}{@{}l@{\hspace{4pt}}c@{\hspace{4.0pt}}ccc@{\hspace{4.0pt}}cccc@{\hspace{4.0pt}}cccc@{\hspace{4.0pt}}cccc@{\hspace{4.0pt}}cccc@{}}
\toprule
  & \multicolumn{1}{c}{$k$ (\%)} & \multicolumn{3}{c}{Size (M)} & \multicolumn{4}{c}{HumanEval} & \multicolumn{4}{c}{HellaSwag} & \multicolumn{4}{c}{GSM8K} & \multicolumn{4}{c}{PPL$\,\downarrow$} \\
\cmidrule(l{2pt}r{2pt}){3-5}\cmidrule(l{2pt}r{2pt}){6-9}\cmidrule(l{2pt}r{2pt}){10-13}\cmidrule(l{2pt}r{2pt}){14-17}\cmidrule(l{2pt}r{2pt}){18-21}
  &   & \multicolumn{1}{c}{$S$} & \multicolumn{1}{c}{$D$} & \multicolumn{1}{c}{$N$} & \multicolumn{1}{c}{none} & \multicolumn{1}{c}{$S$} & \multicolumn{1}{c}{$D$} & \multicolumn{1}{c}{$N$} & \multicolumn{1}{c}{none} & \multicolumn{1}{c}{$S$} & \multicolumn{1}{c}{$D$} & \multicolumn{1}{c}{$N$} & \multicolumn{1}{c}{none} & \multicolumn{1}{c}{$S$} & \multicolumn{1}{c}{$D$} & \multicolumn{1}{c}{$N$} & \multicolumn{1}{c}{none} & \multicolumn{1}{c}{$S$} & \multicolumn{1}{c}{$D$} & \multicolumn{1}{c}{$N$} \\
\midrule
\multicolumn{21}{@{}l}{\textbf{Llama-3.1-8B-Instruct}} \\
\addlinespace[1pt]
\multicolumn{21}{@{}l}{\hspace{0.5em}\emph{Quantization}} \\
\hspace{0.5em}RTN-W4 & 0.5 & 12.6 & 11.1 & 11.1 & 62.2 & 0.6 & 37.8 & 35.4 & 78.2 & 72.4 & 76.0 & 72.8 & 73.0 & 52.0 & 56.0 & 57.0 & 9.0 & 12.0 & 10.8 & 11.2 \\
\hspace{0.5em}AWQ & 0.5 & 16.9 & 6.8 & 6.4 & 57.9 & 0.6 & 45.1 & 48.8 & 78.0 & 71.8 & 77.0 & 78.2 & 74.0 & 52.0 & 65.5 & 68.5 & 8.7 & 12.6 & 9.8 & 9.4 \\
\hspace{0.5em}GPTQ & 0.5 & 14.8 & 8.9 & 8.3 & 57.9 & 1.2 & 37.8 & 42.7 & 77.8 & 73.0 & 77.8 & 76.6 & 78.0 & 52.5 & 58.5 & 59.5 & 8.8 & 12.1 & 10.5 & 9.7 \\
\addlinespace[1pt]
\multicolumn{21}{@{}l}{\hspace{0.5em}\emph{Pruning}} \\
\hspace{0.5em}Magnitude & 0.5 & \textbf{5.3} & 18.4 & 14.9 & 53.0 & 22.6 & 13.4 & 6.1 & 76.8 & 76.4 & 71.8 & 72.4 & 42.5 & 33.5 & 44.5 & 10.0 & 14.4 & 15.3 & 15.6 & 20.7 \\
\hspace{0.5em}Wanda & 0.5 & 16.4 & 7.3 & 6.9 & 63.4 & 0.0 & 47.0 & 45.1 & 78.0 & 73.4 & 77.6 & 77.0 & 78.0 & 47.5 & 72.5 & 71.5 & 8.8 & 12.6 & 10.0 & 9.7 \\
\hspace{0.5em}SparseGPT & 0.5 & 14.8 & 8.9 & 7.9 & 57.9 & 0.0 & 45.1 & 44.5 & 77.4 & 73.8 & 75.8 & 77.2 & 79.0 & 58.0 & 66.5 & 69.5 & 8.8 & 11.6 & 10.3 & 9.8 \\
\addlinespace[1pt]
\multicolumn{21}{@{}l}{\hspace{0.5em}\emph{Training}} \\
\hspace{0.5em}LoRA & 0.25 & 8.1 & 3.0 & 3.1 & 56.1 & 7.3 & 50.0 & 56.1 & 79.0 & 75.6 & 78.4 & 78.2 & 75.5 & 44.0 & 69.5 & 70.5 & 8.2 & 11.4 & 8.6 & 8.8 \\
\addlinespace[1pt]
\midrule
\multicolumn{21}{@{}l}{\textbf{Mistral-7B-Instruct-v0.3}} \\
\addlinespace[1pt]
\multicolumn{21}{@{}l}{\hspace{0.5em}\emph{Quantization}} \\
\hspace{0.5em}RTN-W4 & 0.5 & 13.8 & 10.2 & 10.0 & 35.4 & 8.5 & 23.2 & 33.5 & 81.0 & 80.2 & 80.8 & 80.2 & 44.5 & 40.0 & 42.0 & 43.5 & 6.5 & 7.5 & 6.9 & 6.8 \\
\hspace{0.5em}AWQ & 0.5 & 17.3 & 6.8 & 6.3 & 32.3 & 0.6 & 29.3 & 32.3 & 80.8 & 80.2 & 81.0 & 81.6 & 55.5 & 34.5 & 43.0 & 46.0 & 6.4 & 7.4 & 6.7 & 6.6 \\
\hspace{0.5em}GPTQ & 0.5 & 15.7 & 8.3 & 8.9 & 30.5 & 3.7 & 22.6 & 25.0 & 81.6 & 81.0 & 82.0 & 80.8 & 44.0 & 30.0 & 36.0 & 38.5 & 6.5 & 7.5 & 6.8 & 6.9 \\
\addlinespace[1pt]
\multicolumn{21}{@{}l}{\hspace{0.5em}\emph{Pruning}} \\
\hspace{0.5em}Magnitude & 0.5 & 16.0 & 8.0 & 7.9 & 38.4 & 0.0 & 33.5 & 29.9 & 82.2 & 80.8 & 80.4 & 81.0 & 44.0 & 35.0 & 34.0 & 46.5 & 6.6 & 7.6 & 7.0 & 6.9 \\
\hspace{0.5em}Wanda & 0.5 & 17.3 & 6.7 & 6.0 & 38.4 & 1.2 & 36.0 & 30.5 & 81.6 & 81.0 & 81.8 & 81.4 & 44.5 & 38.5 & 44.0 & 46.5 & 6.4 & 7.4 & 6.7 & 6.6 \\
\hspace{0.5em}SparseGPT & 0.5 & 16.9 & 7.2 & 6.6 & 37.2 & 3.7 & 36.0 & 34.1 & 81.8 & 81.6 & 81.6 & 81.4 & 54.5 & 41.0 & 44.5 & 47.5 & 6.4 & 7.3 & 6.7 & 6.6 \\
\addlinespace[1pt]
\multicolumn{21}{@{}l}{\hspace{0.5em}\emph{Training}} \\
\hspace{0.5em}LoRA & 1 & \textbf{19.6} & 28.7 & 25.0 & 34.1 & 13.4 & 20.1 & 28.0 & 81.4 & 78.6 & 79.8 & 80.4 & 40.0 & 38.0 & 32.5 & 34.5 & 6.4 & 7.0 & 8.1 & 6.8 \\
\addlinespace[1pt]
\midrule
\multicolumn{21}{@{}l}{\textbf{Llama-3.2-3B-Instruct}} \\
\addlinespace[1pt]
\multicolumn{21}{@{}l}{\hspace{0.5em}\emph{Quantization}} \\
\hspace{0.5em}RTN-W4 & 0.5 & \textbf{4.7} & 5.6 & 6.4 & 40.2 & 1.2 & 29.3 & 20.7 & 67.8 & 67.0 & 66.4 & 65.8 & 64.0 & 52.0 & 62.0 & 52.0 & 14.6 & 19.6 & 16.5 & 21.2 \\
\hspace{0.5em}AWQ & 0.5 & 5.8 & 4.5 & 4.5 & 46.3 & 3.7 & 34.1 & 45.1 & 68.8 & 67.0 & 66.8 & 68.4 & 69.0 & 63.5 & 71.5 & 65.5 & 14.1 & 19.4 & 15.8 & 15.8 \\
\hspace{0.5em}GPTQ & 0.5 & \textbf{4.2} & 6.1 & 5.8 & 48.8 & 0.6 & 18.9 & 33.5 & 68.4 & 68.6 & 66.8 & 67.4 & 65.5 & 60.0 & 62.5 & 64.0 & 14.7 & 18.2 & 17.3 & 16.6 \\
\addlinespace[1pt]
\multicolumn{21}{@{}l}{\hspace{0.5em}\emph{Pruning}} \\
\hspace{0.5em}Magnitude & 0.5 & \textbf{4.4} & 5.9 & 6.8 & 36.6 & 0.0 & 25.6 & 9.8 & 67.6 & 63.8 & 64.8 & 63.4 & 68.0 & 56.5 & 59.0 & 44.5 & 16.5 & 21.7 & 19.7 & 25.2 \\
\hspace{0.5em}Wanda & 0.5 & \textbf{4.4} & 5.9 & 5.5 & 38.4 & 2.4 & 13.4 & 37.2 & 68.4 & 67.0 & 66.6 & 67.6 & 66.0 & 60.0 & 64.5 & 61.5 & 14.1 & 18.1 & 16.6 & 16.5 \\
\hspace{0.5em}SparseGPT & 0.5 & \textbf{4.5} & 5.7 & 5.6 & 49.4 & 5.5 & 26.2 & 43.3 & 68.8 & 67.4 & 67.0 & 67.8 & 63.5 & 63.0 & 62.5 & 58.5 & 14.3 & 18.0 & 16.7 & 16.6 \\
\addlinespace[1pt]
\multicolumn{21}{@{}l}{\hspace{0.5em}\emph{Training}} \\
\hspace{0.5em}LoRA & 1 & 11.1 & 5.8 & 5.1 & 40.2 & 1.8 & 34.1 & 39.6 & 70.6 & 69.0 & 68.0 & 70.0 & 60.5 & 49.0 & 59.0 & 58.0 & 12.2 & 16.1 & 13.3 & 13.1 \\
\addlinespace[1pt]
\bottomrule
\end{tabular}
\caption{\textbf{Parameter-partition sizes and ablation results.} Size is reported in millions of parameters. Scores are shown before ablation (none) and after separately ablating $S$, $D$, or $N$ in $M_{\mathrm{opt}}$. Accuracy is reported in percent.}
\label{tab:partition}
\end{table*}

\end{document}